\documentclass{article}
\usepackage{spconf,amsmath,graphicx}
\usepackage{stmaryrd, bbold,xspace,xcolor,url,nicefrac}

\def\Att{\mathcal{A}}
\def\acc{\eta}
\def\hd{D_{\nicefrac{1}{2}}}
\def\dim{n}

\def\ie{\textit{i.e.}\xspace}
\def\eg{\textit{e.g.}\xspace}
\def\etal{\textit{et al.}\xspace}

\def \robic{\texttt{RoBIC}\xspace}

\def \FGSM{\texttt{FGSM}\xspace}
\def \IFGSM{\texttt{I-FGSM}\xspace}
\def \PGD{\texttt{PGD}\xspace}
\def \CW{\texttt{CW}\xspace}
\def \DEEPFOOL{\texttt{DeepFool}\xspace}
\def \BP{\texttt{BP}\xspace}

\def \fgsm{\FGSM~\cite{Goodfellow:2015aa}\xspace}
\def \ifgsm{\IFGSM~\cite{Kurakin:2017aa}\xspace}
\def \pgd{\PGD~\cite{Kurakin:2017aa}\xspace}
\def \cw{\CW~\cite{carlini2017towards}\xspace}
\def \deepfool{\DEEPFOOL~\cite{7780651}\xspace}
\def \bp{\BP~\cite{Zhang:2021aa}\xspace}

\def \SURFREE{\texttt{SurFree}\xspace}
\def \RAYS{\texttt{RayS}\xspace}
\def \GEODA{\texttt{GeoDA}\xspace}
\def \QEBA{\texttt{QEBA}\xspace}

\def \surfree{\SURFREE~\cite{Maho:2020aa}\xspace}
\def \rays{\RAYS~\cite{Chen:2020ab}\xspace}
\def \geoda{\GEODA~\cite{Rahmati:2020aa}\xspace}
\def \qeba{\QEBA~\cite{Li:2020aa}\xspace}

\def \ares{\texttt{ARES}~\cite{Dong:2020aa}\xspace}
\def \robustbench{\texttt{RobustBench}~\cite{Croce:2020aa}\xspace}
\def \robustvision{\texttt{RobustVision}~\cite{Lab:aa}\xspace}
\def \robustml{\texttt{RobustML}\xspace}
\def \adbd{\texttt{ADBD}~\cite{Chen:2020ab}\xspace}

\title{\robic: A benchmark suite for assessing classifiers robustness}
\name{Thibault Maho, Beno\^it Bonnet, Teddy Furon$^\dagger$\thanks{$^\dagger$Thanks to ANR and AID for funding Chaire IA \textit{SAIDA} (ANR\_20-CHIA-0011-01).} and Erwan Le Merrer}
\address{Univ. Rennes, Inria, CNRS, IRISA, Rennes France}
\begin{document}
%
\maketitle

\begin{abstract}
Many defenses have emerged with the development of adversarial
attacks. Models must be objectively evaluated accordingly. This paper
systematically tackles this concern by proposing a new parameter-free
benchmark we coin \robic. \robic fairly evaluates the robustness of image classifiers
using a new \textit{half-distortion} measure.
It gauges the robustness of the network against white and black box attacks, independently of its accuracy.
\robic is faster than the other available benchmarks.
We present the significant differences in the robustness of 16 recent models as assessed by \robic.

We make this benchmark publicly available for use and contribution at \url{https://gitlab.inria.fr/tmaho/robustness_benchmark}.

\end{abstract}
%
\section{Introduction}
\label{sec:intro}
Deep learning models are vulnerable to adversarial perturbations.
This is especially true in image classification in computer vision.
This weakness is unfortunatley undermining the developement of
`Artificial Intelligence'. In particular, adversarial attacks are a
serious threat for security oriented applications. Attackers willing
to bypass security countermeasures might use the deep learning models
as the weakest link. A deluge of research papers now propose defenses
to block such an attacker, and adaptive attacks against these
defenses. This is an endless arms race, and systematic benchmarks to
evaluate the state of the threat are greatly required.



It is currently extremely difficult to have a clear view on what is
truly working in this domain. The cliché is that no two papers report
the same statistics for the same attack against the same model over
the same image set.  This is mostly due to that an attack is an
algorithm with many parameters.
Its power 
is indeed highly dependent of
these parameters.  These values are rarely specified in research
papers.

There exist benchmarks in the litterature, such as \ares, \robustbench,
\robustvision, \adbd. They aim at providing a
better understanding of the robustness of image classifiers.
Yet, they fall short because their slowness prevents them from tackling large image dataset like ImageNet. 
They only operate on CIFAR-10 
or MNIST. Also, they resort to attacks which are not all state-of-the-art.

This paper proposes \robic, to consider these concerns and develop a
benchmark tool to measure the robustness of image classifiers in a
modern setup.


\section{Difficulties}
\label{sec:difficulties}
This sections explains the difficulties for setting up a benchmark measuring the robustness of image classifiers.

\subsection{Notation}
An attack is a process forging an image $I_a = \Att(I_o, M, \Pi)$, where $I_o$ is the original image, $M$ is the target model, and $\Pi$ is a set of attack parameters.
The ground truth label of $I_o$ is denoted by $y_o$.
The boolean function $\mathbb{1}(I_a,y_o) = [M(I_a)\neq y_o]$ tells whether the attack deludes classifier $M$ in the untargeted attack scenario: the prediction $M(I_a)$ is not the ground truth.
The distortion between $I_o$ and $I_a$ is denoted by $d(I_a,I_o)$.

Some statistics like the probability of success and the average distortion are extracted from the adversarial images forged from the test set.
They depend on the attack $\Att$ and its set of parameters $\Pi$.
Therefore, it can not play the role of a measure of robustness of a given model.
The first difficulty is to get rid off the impact of parameters $\Pi$.

\subsection{The best effort mode}
The parameters $\Pi$ have a huge impact on the power of an attack.
For instance, some attacks like \fgsm, \ifgsm, \pgd are distortion constrained in the sense that $\Pi$ is strongly connected to a distortion budget.
If this  budget is small, the probability of the success of the attack is small. If it is large, this probability is close to 1 but the distortion is too big.
Hence, it is hard to find the best setting to make these attacks competitive.
Our strategy, so-called `best effort mode', reveals the intrinsic power of an attack by finding the best setting for any image: $I_a = \Att(I_o, M, \Pi^\star)$ with
\begin{equation}
\Pi^\star = \arg\min_{\Pi:\mathbb{1}(\Att(I_o, M, \Pi),y_{o})=1} d(\Att(I_o, M, \Pi),I_o).
\end{equation}
The best effort mode makes the measurement of the robustness independent from an arbitrary global setting $\Pi$.
Yet, it is costly in terms of computations. Attacks with few parameters are preferred since the search space is smaller.

\subsection{Worst case attacks}
A second difficulty is to make the robustness score independent of the attack.
Ideally, we would like to know the worst case attack to certify the robustness of a model. 
An option proposed by benchmarks \robustvision and \ares is to consider a set of $J = 11$ attacks as outlined in table \ref{table:benchmarks_comparison}.
This is again costly as each image of the test set has to be attacked $J$ times.
Yet, a benchmark happens to be useful if it is fast enough so to assess the robustness of many models.
The best effort mode over an ensemble of attacks is out of reach.
This is the reason why we need to focus on fast worst case attacks in the sense that they achieve their best effort mode within limited complexity.
Section \ref{sec:fast} focuses on these attacks.

\subsection{The choice of the metric}
The game between attack $\Att$ and model $M$ over the test set is summarized by the operating characteristic $D\to P(D)$ relating the distortion $D$ and the probability of success $P(D)$:
\begin{equation}
	P(D) := n^{-1}\sum_{i: d(I_{a,i},I_{o,i})\leq D }^{n} \mathbb{1}(I_{a,i},y_{o,i}).
\end{equation}
In other words, $P(D)$ is the fraction of images that the attack succeeded to hack within a distortion budget $D$. 
Many benchmarks gauge the robustness by $P(D_b)$ at an arbitrary distortion $D_b$: \eg\
\robustbench score is  $P(D = 0.5)$.
This measure is pointwise and dependent on $\acc(0)$.


\section{The benchmark}
\label{sec:benchmark}
This section justifies the recommendations made in our benchmark and defines the measure of robustness.

\textbf{Pixel domain.} Our benchmark is dedicated to image classification.
As a consequence, the distortion is defined on the pixel domain:
An image $I$ is defined in the space $\llbracket0,255\rrbracket^n$ with $n=3RC$ pixels for 3 color channels, $R$ rows and $L$ columns.
Most papers in the field measure distortion after the transformation of the image in a tensor $x\in\mathcal{X}^n$.
This is a mistake preventing a fair comparison: for most models $\mathcal{X}=[0,1]$, but for some others $\mathcal{X}=[-1,1]$ or $\mathcal{X}=[-3,3]$.

We outline that an adversarial image is above all an image, \ie\ a discrete object $I_a\in\llbracket0,255\rrbracket^n$.
Again, most attacks output a continuous tensor $x_a\in\mathcal{X}^n$, neglecting the quantization.
This is a mistake: in real-life, the attacker has no access to $x_a$, which is an auxiliary data internal of the model.

\textbf{Distortion.} The distortion is defined as the root mean square error: $d(I_a,I_o) := \|I_a - I_o\|_2 / \sqrt{n}$. 
This is easily interpretable: if $I_{a,i} = I_{o,i} \pm \epsilon$, $\forall i\in\llbracket 1,n\rrbracket$, then $d(I_o,I_a)=\epsilon$.
It is easily translated into a PNSR as image processing professionals do: $\mathrm{PSNR} = 48.13 - 20\log_{10}(d(I_o,I_a))$ dB.
Adversarial perturbations usually spread all over the image and have small amplitude like in invisible watermarking.
This is a case where measures based on $\ell_2$ norm remain good indicators of the quality.  
A perceptual similarity is obviously better, but more complex and less interpretable.
 
\textbf{Test set.} The input of the model is a natural and large image. Assessing the robustness of models on specific dataset like MNIST (almost black and white), or on tiny images like CIFAR does not reflect the complexity of the problem. Our benchmark considers natural images of at least $224\times 224$ pixels as provided in ImageNet.

\textbf{Measure of robustness.} 
\label{subsec:half_distortion}
Let us define the accuracy function $\acc(D) := 1-P(D)$. 
The value $\eta(0)$ is the classical accuracy of the model over original images. 
Function $\eta(D)$ is by construction non increasing and should converge to 0 as the distortion $D$ increases.
After observing many accuracy functions $\acc$ for different models and attacks, we notice that they share the same prototype:
\begin{equation}
\acc(D) = \acc(0)~e^{-\lambda D} \quad \text{with }  \lambda \in \mathbb{R^+}.
\end{equation}
Like in nuclear physics, we define the half-distortion $\hd$ as the distortion needed to reduce to half the initial accuracy:
\begin{equation}
\label{eq:HD}
\acc(\hd) = \acc(0)/2,\quad \hd = \lambda^{-1}\log(2).
\end{equation}

This approximation is verified experimentally with an average coefficient of determination $R^2$ of 99\%.
The half-distortion $\hd$ will be the keystone of the proposed metric of robustness.
A model is then characterized by three separated concepts: its generalization ability $\acc(0)$ and its robustnesses $\hd$ against black-box and white-box attacks.


\section{Fast Attacks}
\label{sec:fast}
The recent trend in adversarial examples is to design fast attacks with state-of-the-art performances.

\subsection{Fast black-box attacks}
In the black-box decision based setup, the attacker can query  a model and observes the predicted class.
The complexity of the attack is gauged by the number of queries $K$ needed to find an adversarial image of low distortion.   

There has been a huge improvement on the amount of queries recently.
Brendel \etal report in the order of one million of queries for one image in one of the first decision based black-box  \texttt{BA}~\cite[Fig.~6]{Brendel:2018aa}.
Then, the order of magnitude went down to tens of thousands~\cite[Fig.~4]{Chen:2020aa} \cite[Fig.~5]{Li:2020aa} and even some thousands in~\cite[Fig.~2]{Rahmati:2020aa}.
Current benchmarks use others black-box attacks, which are either decision-based (\texttt{Square Attack}~\cite{andriushchenko2020square} in \robustbench is score-based), or not state-of-the-art (like Gaussian noise in \robustvision,  or \texttt{BA}~\cite{Brendel:2018aa} in \ares).

\surfree and \rays are the only \emph{decision-based} papers with less than one thousand of calls on ImageNet.
Yet, \rays is designed to minimize the $\ell_\infty$ distortion, whereas $\surfree$ targets $\ell_2$.
Sect.~\ref{sec:Experiments} investigates which attack is the best candidate for a fast benchmark.

\subsection{Fast white-box attacks}
In the white-box setup, the attacker can compute a loss function and its gradient thanks to auto-differentiation and back-propagation.
The complexity is usually gauged by the number of gradient computations.
Current benchmarks use different white-box attacks: \robustbench relies on \pgd (with 2 parameters $\Pi$), \robustvision use \deepfool, and \ares \cw.

Again, the need for powerful but fast attacks is of utmost importance for a practical benchmark.
A promising attack is \bp designed for low complexity budget.
Its first stage finds an adversarial example as quickly as possible.
It is nothing more than a gradient descent of the loss $L$ with acceleration.
At iteration $t+1$:
\begin{equation}
\label{eq:BP}
I_a^{(t+1)} = I_a^{(t)} - \alpha\gamma(t+1) \eta\left(\nabla L(I_a^{(t)})\right),
\end{equation}  
where $I_a^{(0)} = I_o$, $\eta(x) = x/\|x\|_2$, and $\gamma(t)$ is a series of increasing values, hence the acceleration.
Stage 1 finishes when $I_a^{(t+1)}$ becomes adversarial.
Stage 2 aims at lowering the distortion while maintaining the image adversarial (see~\cite{Zhang:2021aa}).

We develop a variant to aggressively downsize the number of gradient computations.
Parameter $\alpha$ is heuristically set up to $0.03$ in~\cite{Zhang:2021aa}.
This value is certainly too big for images close to the class boundary and too small for those further away.
One costly option is the best effort mode which finds the best $\alpha$ thanks to a line search (see Sect.~\ref{sec:difficulties}).
We propose the following simple method inspired by \deepfool.
When applying~\eqref{eq:BP} to the first order approximation of the loss:
\begin{equation}
L(I_o+p) \approx L(I_o) + p^\top \nabla L(I_o),\\
\end{equation}
then $\eta\left(\nabla L(I_a^{(t)})\right)=\eta(\nabla L(I_o))$ and \texttt{BP} cancels the loss for
\begin{equation}
\label{eq:Alpha}
\alpha = \frac{L(I_o)}{\|\nabla L(I_o)\|_2\sum_{j=1}^\kappa\gamma(k) }
\end{equation}
within $\kappa$ iterations.
We fix $\kappa = \lfloor K/3\rfloor$ where $K$ is the total iteration budget encompassing stages 1 and 2.

Sect.~\ref{sec:Experiments} compares these attacks to identify the worst case.

\subsection{Quantization}
The adversarial samples are quantified in the pixel domain to create images.
The first option considers the quantization as a post-processing not interfering with the attack.
The second option performs quantization at the end of any iteration.
These options are tested on several black and white box attacks.
The quantization will be a post-processing for white-box attacks as recommended in~\cite{10.1145/3369412.3395062},
whereas the second option give better results on black-box attacks.

\section{Experiments}
\label{sec:Experiments}
All the attacks are run on 1000 ImageNet images from the ILSVRC2012's validation set with size $\dim = 3 \times 224 \times 224$.

\subsection{Selecting the worst case attacks}
\textbf{Black box attacks}:
Figure~\ref{fig:distortion_evolution_BB} compares the evolution of the half-distortion~\eqref{eq:HD} in function of the query amount for four decision-based black-box attacks: \surfree, \rays, \geoda, and \qeba.
\SURFREE and \RAYS reach their best effort within 3000 queries, while \QEBA and \GEODA do not since their $\hd$ still decrease after 5000 queries.
Yet, \SURFREE obtains quantified adversarials with much lower distortion. Therefore, our benchmark only needs this attack.
The number of queries is kept at 5000 to be sure to reach the optimal value of $D_{1/2}$.


\begin{figure}[t]
	\centering
	\includegraphics[width=0.7\columnwidth]{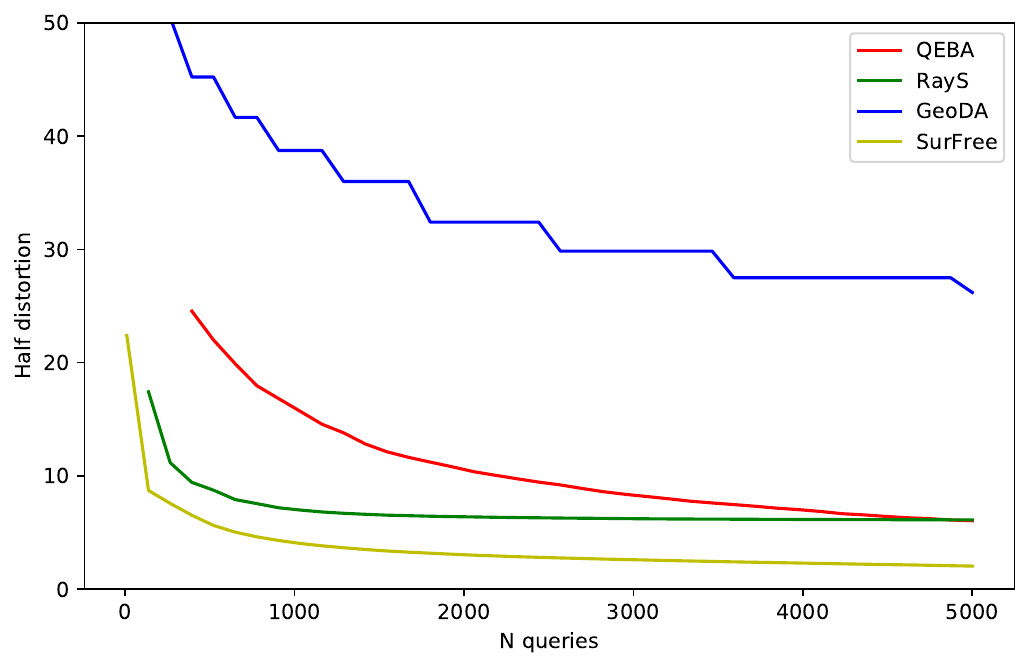}
	\caption{Evolution of $\hd$ with the complexity budget for black box setup. Attacks on EfficientNet~\cite{pmlr-v97-tan19a}}
	\label{fig:distortion_evolution_BB}
\end{figure}

\begin{figure}[t]
	\centering
	\includegraphics[width=0.7\columnwidth]{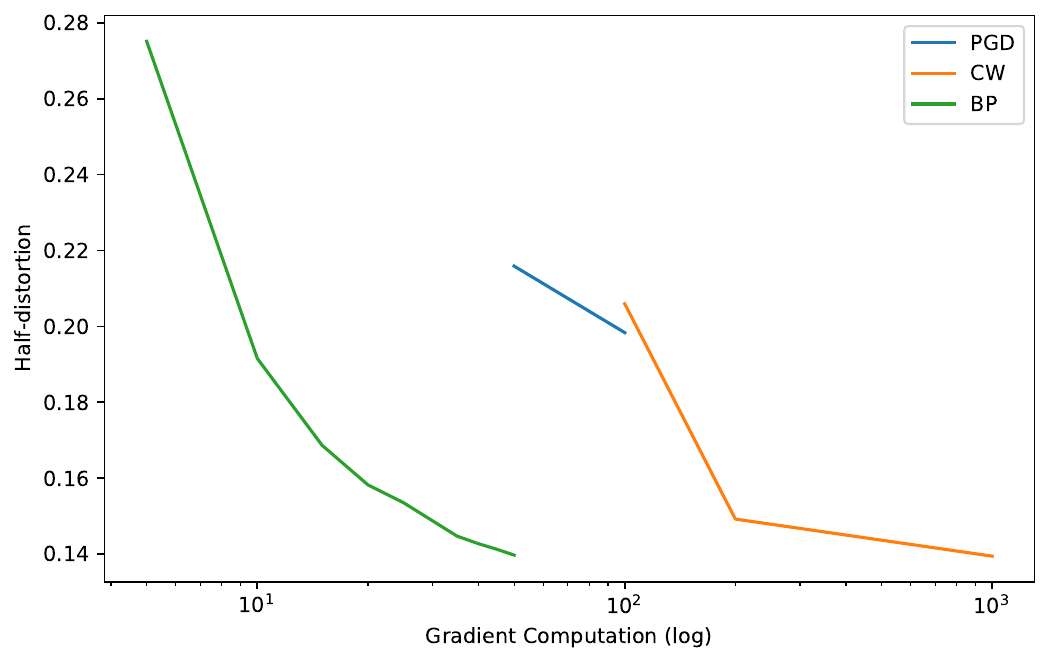}
	\caption{Evolution of $\hd$ with the complexity budget for white box setup. Attacks on EfficientNet~\cite{pmlr-v97-tan19a}}
	\label{fig:distortion_evolution_WB}
\end{figure}

\textbf{White box attacks}: Figure~\ref{fig:distortion_evolution_WB} compares three white-box-attacks in the best effort mode: \pgd, \cw, and \bp with our trick~\eqref{eq:Alpha}. They all reach the same $\hd$ when given a large complexity budget.
Yet, \BP converges faster than the others.
Our benchmark uses this version of \BP to evaluate the white-box-robustness.

\begin{table}[b]
	\centering
	\resizebox{\linewidth}{!}{%
		\setlength\tabcolsep{2pt}
		\begin{tabular}{|c|c|c|c|c|}
			\hline
			Benchmark & Domain & Nb. attacks & Measures & Runtime \\
			\hline
			\robic & $\llbracket 0, 255 \rrbracket ^n $  & 1 WB + 1 BB & Half-distortion $\ell_2$ & 43s \\
			\hline
			\robustbench & $[0, 1]^n$ & 3 WB + 1 BB & Success-Rate for & 48s \\
			& & & fixed budget ($\ell_2$ or $\ell_{\infty}$) & \\
			\hline
			\adbd & $[0, 1]^n$ & 1 BB & Distance $\ell_\infty$ & 360s\\
			\hline
			\robustvision & $[0, 1]^n$ & 6 WB + 5 BB & Median Distance $\ell_2$ & 200s\\
			\hline
			\ares & $[0, 1]^n$ & 5 WB + 10 BB & Success-Rate vs Budget  & Too long\\ 
			& & & ($\ell_2$, $\ell_{\infty}$ or queries) & \\
			\hline
		\end{tabular}
	}
	\caption{Benchmarks Comparison. Average Runtimes per ImageNet Image with ResNet50~\cite{madry2018towards}.}
	\label{table:benchmarks_comparison}
\end{table}

\subsection{Comparison with other benchmarks}
Table~\ref{table:benchmarks_comparison} lists several benchmarks.
Most of them evaluate the robustness as the success-rate under a prescribed $\ell_2$ or $\ell_{\infty}$ distortion budget.
But, these budgets are set arbitrarily or even not constant within the same benchmark for \robustml.
Our half-distortion~\eqref{eq:HD} is parameter-free. It returns an accurate, reliable  and fair measurement of robustness.

Some benchmarks need many attacks to get a full vision of the robustness: \ares and \robustvision use 11 attacks.
This is too time-consuming. 
On the contrary, \adbd focuses on a single black-box attack, which is indeed outdated.
\robustbench condenses four attacks in one measure elegantly: for a given image,
if the first simple attack does not succeed within the distortion budget, then the second more complex one is launched \textit{etc.}
The total runtime heavily depends on the distortion budget.
Yet, black-box and white-box attacks use different mechanisms.
Our benchmark reports a measurement for each separately.

\begin{table}[t]
	\centering
	\resizebox{\linewidth}{!}{%
		\setlength\tabcolsep{2pt}
		\begin{tabular}{|c|r||c|c|c|}
			\hline
			Model & Parameters & Accuracy & \multicolumn{2}{|c|}{$\hd$ } \\
			& (millions) & $\acc(0)$ & white box & black box \\
			\hline
			AlexNet~\cite{DBLP:journals/corr/Krizhevsky14} & 62.38 & 56.8 & 0.19 & 2.17 \\
			CSPResNeXt50~\cite{wang2020cspnet} & 20.57 & 84.6 & 0.13 & 4.48 \\
			DualPathNetworks 68b ~\cite{10.5555/3294996.3295200} & 12.61 & 83.8 & 0.08 & 3.82 \\
			MixNet Large~\cite{2019arXiv190709595T} & 7.33 & 84.2 & 0.12 & 2.96 \\
			MobileNetV2~\cite{Sandler_2018_CVPR} & 5.83 & 80.1 & 0.09 & 2.90 \\
			ReXNet 200~\cite{2020arXiv200700992H} & 16.37 & 85.4 & 0.14 & 3.89 \\
			RegNetY 032~\cite{Radosavovic2020} & 19.44 & 85.8 & 0.11 & 4.94 \\
			SEResNeXt50 32x4d~\cite{hu2018senet} & 27.56 & \bf 85.9 & 0.12 & \bf 5.01 \\
			VGG16~\cite{DBLP:journals/corr/SimonyanZ14a} & 138.00 & 74.9 & 0.09 & 2.44 \\
			\hline
			EfficientNet AdvProp~\cite{Xie_2020_CVPR_advprop} & 5.29 & 84.3 & \bf 0.31 & 4.35 \\
			EfficientNet EdgeTPU Small~\cite{pmlr-v97-tan19a} & 5.44 & 82.8 & 0.15 & 3.16 \\
			EfficientNet NoisyStudent~\cite{Xie_2020_CVPR_ns} & 5.29 & 82.7 & 0.19 & 2.37 \\
			EfficientNet~\cite{pmlr-v97-tan19a} & 5.29 & 82.8 & 0.17 & 3.56 \\
			\hline
			ResNet50 (torchvision)~\cite{He_2016_CVPR}  & 25.56 & 77.9 & 0.10 & 2.77 \\
			ResNet50 (timm)~\cite{He_2016_CVPR}  & 25.56 & 80.5 & 0.15 & 4.35 \\
			ResNet50 AdvTrain~\cite{madry2018towards} & 25.56 & 60.8 & \bf 2.56 & \bf 9.88 \\
			\hline
		\end{tabular}
	}
	\caption{Benchmark of models with 1.000 ImageNet Images}
	\label{table:our_benchmark_results}
\end{table}

\subsection{Benchmarking models}
Table~\ref{table:our_benchmark_results} compares standard models from
\textit{timm}~\cite{rw2019timm} and \textit{torchvision}~\cite{10.1145/1873951.1874254} libraries.
Here are some intriguing results.


\textbf{Robustness in white box \textit{vs.} black box}.
One does not imply the other.
Fig.~\ref{fig:half_distortion_plot} even shows a negative correlation.
However, some models escape this rule. For instance, VGG16 is neither robust in black box nor in white box.
EfficientNet AdvProp~\cite{Xie_2020_CVPR_advprop} follows the opposite trend.
We believe that black-box robustness reveals the complexity of the borders between classes, and white-box robustness
indicates how close natural images are from the borders.
This highlights the importance of having two different measurements.

\textbf{The importance of the training procedure.}
There is on average a factor 20 between the half-distortions in white and black box.
This factor drops to 4 and 10 for the models adversarially trained:
ResNet50~\cite{madry2018towards}, EfficientNet AdvProp~\cite{Xie_2020_CVPR_advprop}. 

Table~\ref{table:our_benchmark_results} lists four EfficientNet models sharing the same architecture but different training procedures.
Their accuracies are similar but there is up to a factor of 2 between the robustnesses.
The same holds on the three variants of Resnet50.
The gaps in accuracy and robustness are noticeable with standard models from \textit{timm}~\cite{rw2019timm} and \textit{torchvision}~\cite{10.1145/1873951.1874254}. It is even more visible with adversarial training from \cite{madry2018towards}: the gain in robustness is impressive but at the cost of a big drop in accuracy.



\begin{figure}[t]
	\centering
	\includegraphics[width = 0.9 \columnwidth]{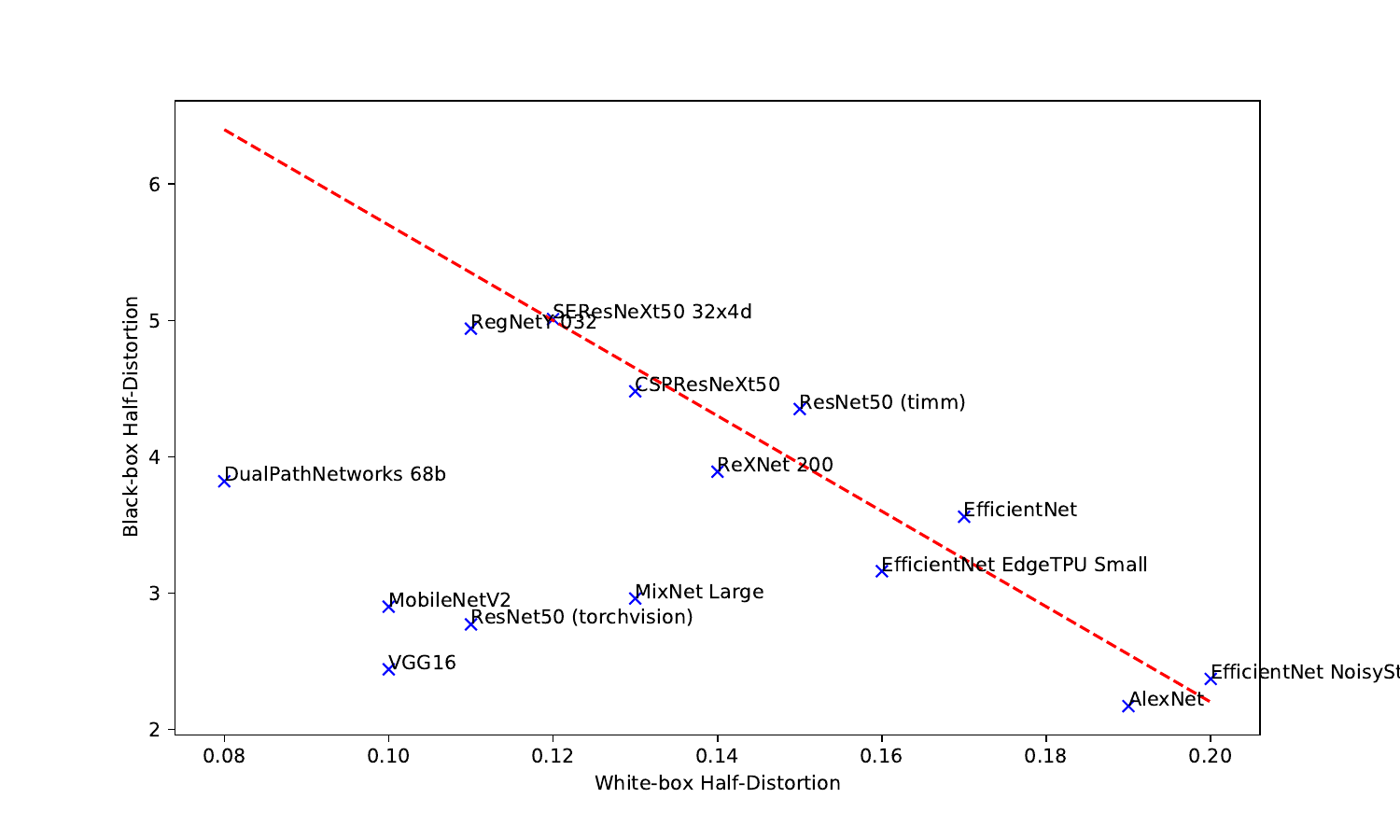} 
	\caption{Black-box $\hd$ as a function of white-box $\hd$.}
	\label{fig:half_distortion_plot}
\end{figure}
 

\section{Conclusion}

The paper introduces a rigorous benchmark based on a new and independent measurement of robustness: the half distortion.
\robic is faster than the other benchmarks. This allows to tackle larger images which is more realistic.

In addition to the accuracy, \robic gives the black box robustness, and white box robustness.
We believe that the first indicates how far away the class boundaries lie from the images whereas the last reflects how curved are the boundaries. 
As the other benchmarks, two limitations hold: The network must be differentiable to run a white box attack,
and deterministic to run a black box attack.



\bibliographystyle{IEEEbib}
\bibliography{icip}

\end{document}